\documentclass{article}

\usepackage{microtype}
\usepackage{graphicx}
\usepackage{subfig}
\usepackage{booktabs} 
\usepackage{enumerate}
\usepackage{multicol}

\usepackage{hyperref}

\usepackage[accepted]{icml2019}

\icmltitlerunning{GAN-GAN}

\begin{document}

\twocolumn[
\icmltitle{GAN You Do the GAN GAN?}



\icmlsetsymbol{equal}{*}

\begin{icmlauthorlist}
\icmlauthor{Joseph Suarez}{}
\end{icmlauthorlist}

\icmlkeywords{Machine Learning, ICML}

\vskip 0.3in
]

\begin{abstract}
Generative Adversarial Networks (GANs) have become a dominant class of generative models. In recent years, GAN variants have yielded especially impressive results in the synthesis of a variety of forms of data. Examples include compelling natural and artistic images, textures, musical sequences, and 3D object files. However, one obvious synthesis candidate is missing. In this work, we answer one of deep learning's most pressing questions: GAN you do the GAN GAN? That is, is it possible to train a GAN to model a distribution of GANs? We release the full source code for this project under the MIT license.\footnote{\textbf{Full source code: \newline https://github.com/jsuarez5341/gan-you-do-the-gan-gan}}

\end{abstract}

\section{Introduction, Background, Related Work}
GANs \cite{2014arXiv1406.2661G} have become perhaps the single most prevalent class of generative models in recent years, spawning hundreds of variants and an entire subfield of surrounding work. While they are notoriously difficult to train and often suffer from mode collapse, under the right conditions, GANs have been shown to generate compelling artificial data distributions and are perhaps best known for realistic image synthesis \cite{2017arXiv170310593Z}.

Our objective is to apply GANs to a different class of data: GANs themselves. Instead of using a GAN to model images, we use a GAN to model a distribution of GANs that model images. We refer to this architecture over architectures as a GAN-GAN. We do not consider GAN-GAN-GANs or GAN-GAN-GAN-GANs in this work. While these are straightforward to implement, they would require an exponential parameter budget (at least as formulated in the present work) which we lack the hardware to support.

Hypernetworks \cite{2016arXiv160909106H} and extensions thereof \cite{2018arXiv180101952D, NIPS2017_6919} have also explored the concept of using one network to generate the weights of another. However, to our knowledge, all such settings typically learn both the architecture and the meta-architecture end-to-end. In contrast, we are interested in whether it is possible to directly learn a distribution over architectures within the standard setting of generative modeling. We treat GANs themselves as examples and learn a GAN-GAN by training on small set of trained GANs.

This paper is a joke; however, the results are in fact real. Interpretability is a key problem in deep learning, especially in models to be deployed in real world systems. Generative modeling over networks could serve as a useful tool for visualization and analysis of network decisions and results.

\begin{figure}[t!]
\begin{center}
\resizebox{\linewidth}{!}{
\includegraphics{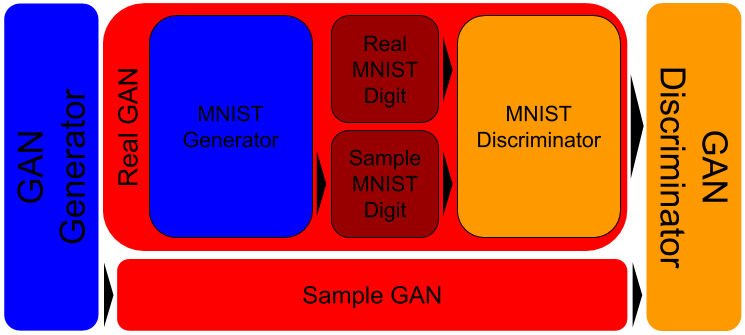}
}
\end{center}
\vspace{-5mm}
\caption{GAN-GAN: a GAN trained on a dataset of GANs}\label{fig:cent}
\label{fig:environment}
\vspace{-5mm}
\end{figure}

\begin{algorithm}[b!]
   \caption{GAN-GAN Training. We first train a set of GANs and save snapshots of the parameters each epoch. We then train a GAN-GAN (a GAN over GANS) by treating each snapshot as an individual training example.}
   \label{alg:pipeline}
\begin{algorithmic}
   \FOR{GAN Index = 1...\#Networks}
      \STATE Initialize an MNIST GAN
      \FOR{Epoch = 1...\#Epochs}
         \STATE Train the MNIST GAN for one epoch
         \STATE Save a snapshot of the GAN parameters
      \ENDFOR
   \ENDFOR
   \STATE Load all snapshots of all GANs into a dataset with \#Networks $\times$ \#Epochs examples
   \STATE Train a GAN over the dataset of GAN snapshots
\end{algorithmic}
\end{algorithm}

\section{Methods}

GANs formulate a two player game between networks. Formally, GANs defines a Generator $\mathcal{G}$ and a Discriminator $\mathcal{D}$. The Discriminator models the probability $P(\mathcal{G}|x)$ that a given example $x$ is fake. The Generator maximizes the probability $P(\mathcal{D}(\mathcal{G}(z)))$ ($z$ is sampled noise) that the Discriminator will output an incorrect prediction. A GAN-GAN is simply a GAN trained on a dataset of GAN weights.

\begin{figure*}
\hspace{-0.5cm}\makebox[\textwidth][c]{\includegraphics[width=1.1\linewidth]{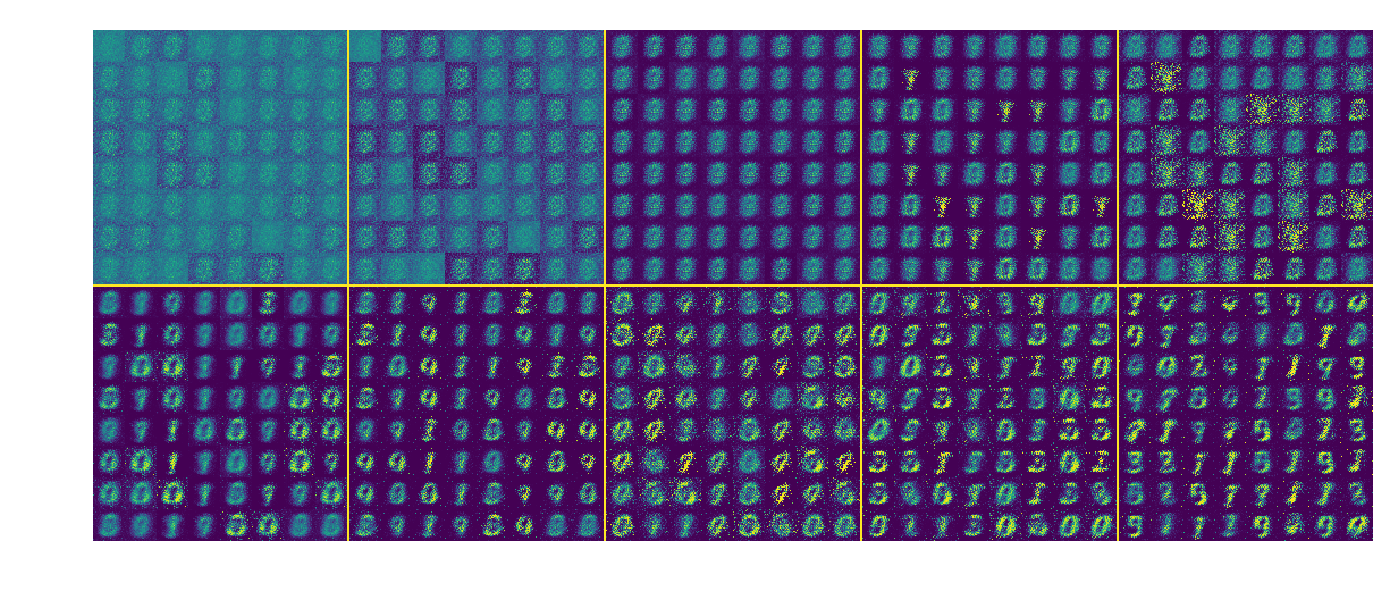}}
\vspace{-10mm}
\begin{center}
\caption{Example samples from the training of an MNIST GAN (top-bottom left-right: epochs 1, 2, 10, 25, 27, 30, 32, 35, 40, 49)}\label{fig:fent}
\end{center}
\hspace{-0.5cm}\makebox[\textwidth][c]{\includegraphics[width=1.1\linewidth]{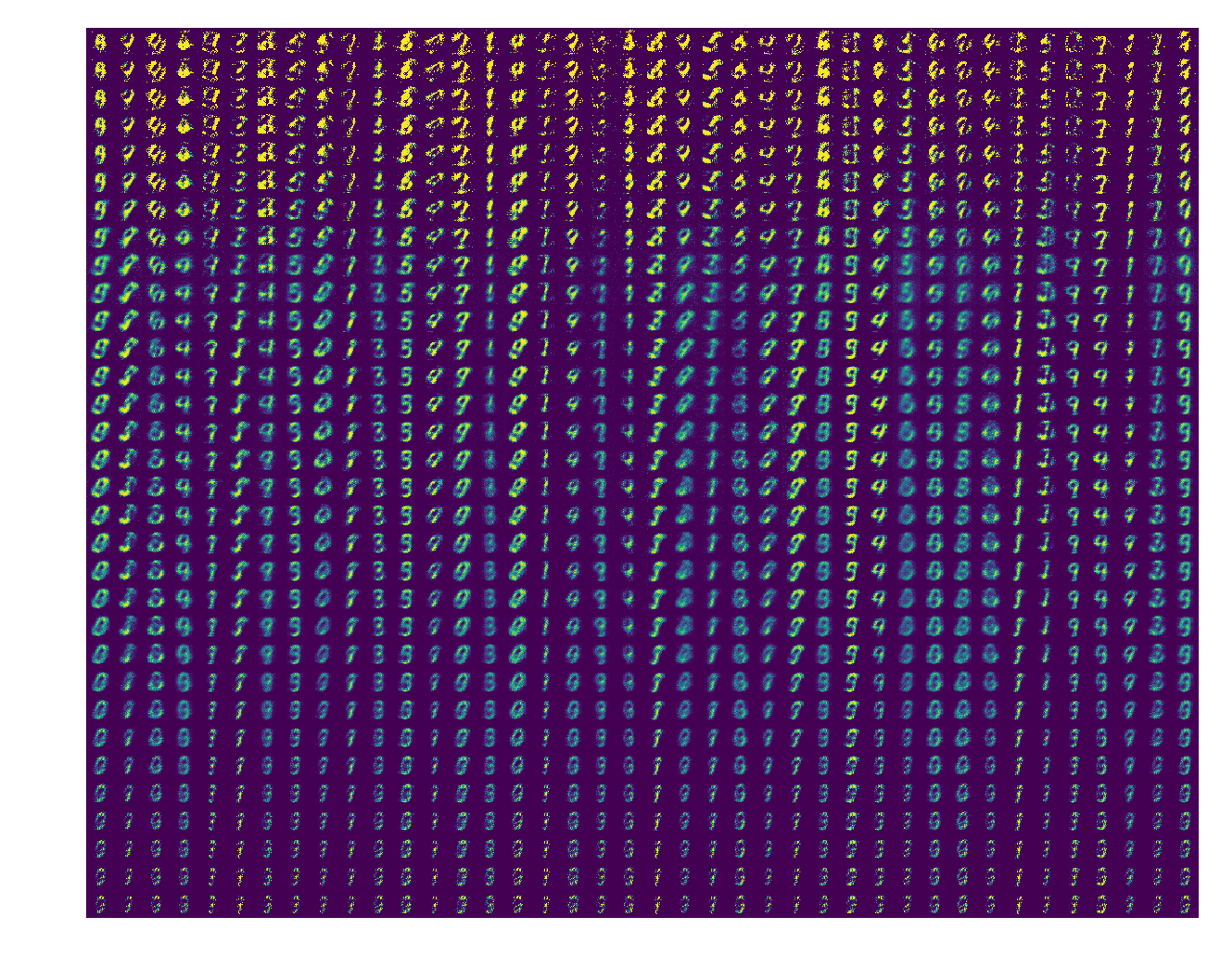}}
\vspace{-10mm}
\begin{center}
\caption{Image samples from GANs sampled from the trained GAN-GAN. Rows correspond to GANs linearly sampled from 1D GAN-GAN latent space in the interval (-2, 2). Columns correspond to a particular noise vector input to all GANs.}\label{fig:fent}
\end{center}
\label{fig:explore}
\end{figure*}

\newpage

\section{Experiments and Discussion}

\paragraph{GAN Architecture}
The generator and discriminator are three layer (input-hidden-output) fully connected neural networks with hidden dimension 64. We use Leaky ReLU\cite{2015arXiv150500853X} activations with 0.2 negative slope after the input and hidden layers. We use \textit{tanh} for the generator output (dimensionality $28 \times 28 = 784$ to match MNIST) and \textit{sigmoid} for the discriminator output (dimensionality 1). The generator samples from latent dimension 64.

\paragraph{GAN-GAN Architecture}
The GAN-GAN generator and discriminator have the same layer and activation structure as the MNIST GAN. The input dimensionality is 113745, which is equal to the dimensionality of the GAN parameter vector. We found that using a smaller hidden dimension for the discriminator (8) than the generator (64) helped to stabilize training. We use latent dimension 1 for the GAN-GAN in order to enable visualizations. Results improve with a larger latent space.

\paragraph{Training}
We use Adam\cite{2014arXiv1412.6980K} for all networks. The learning rate is fixed to 0.0002; all other parameters are PyTorch defaults. We use batch sizes 128 and 32 for MNIST and the GAN-GAN, respectively. As described in Algorithm 1, we train 35 MNIST GANs for 100 epochs each, saving snapshots of the weights at each epoch. We train the GAN-GAN for 250 epochs using these 3500 snapshots as training examples. 

\paragraph{Results}
In order to evaluate the performance of the GAN-GAN, we first linearly sample 32 GANs from the 1-dimensional latent space of the GAN-GAN. We then fix 40 noise samples with the same dimensionality as the GAN latent space. Finally, we sample 40 images from each GAN using the fixed noise samples. Fig. 3 shows the results. The GAN-GAN produces neatly ordered GAN samples according to image quality. Surprisingly, the GAN-GAN actually exhibits better performance than the trained GANs. Given that high quality GANs appear more real than low quality GANs, it is possible that the GAN-GAN learned to bootstrap its own performance by implicitly combining snapshots \cite{2018arXiv180700847L} in latent space, in effect producing a smoothed history over training.

\textbf{Discussion}

While the GAN-GAN samples are of high quality with respect to the MNIST GAN samples, the MNIST GAN samples themselves are of fairly low quality (Fig. 2). As the weight dimensionality of each GAN is equal to the input dimensionality of the GAN-GAN, we were constrained to use a very small network for the MNIST GAN. One obvious solution is to use a more parameter efficient network architecture such as DCGAN \cite{2015arXiv151106434R}. While we were able to obtain significantly better image samples using a slightly modified DCGAN with under 30k parameters, we found it more difficult for the GAN-GAN to learn to model DCGAN with very small amounts of data. This is expected, as small perturbations to the weights in the first layers of the significantly deeper DCGAN model can significantly alter the output of the final layer. While it is likely that our approach would be viable with a larger dataset of DCGAN snapshots, this work was conducted with limited personal hardware. As such, we opted to use a small fully connected model and were unable to test GAN-GANs over DCGAN variants in detail.

\section{Conclusion}
We train a GAN over training snapshots of small, fully connected MNIST GANs and demonstrate that generative models are capable of learning useful representations of other generative models. In short: GAN you do the GAN GAN? Yes you GAN!

{\small
\bibliography{example_paper}}
\bibliographystyle{icml2019}

\end{document}